\documentclass{article}
\usepackage{graphicx}
\usepackage{graphics}
\usepackage{amsfonts}
\usepackage{marvosym}
\newcommand{\R}{\mathbb{R}}

\begin{document}
\title{\huge{\em Sixth Annual Boole Lecture in Informatics}
\\ \  \\    
The Correspondence Analysis Platform for Uncovering Deep
Structure in Data and Information 
}

\author{Fionn Murtagh \\
Science Foundation Ireland, Wilton Place, Dublin 2, Ireland; \\
and Department of Computer Science, Royal Holloway, University of \\ 
London, Egham TW20 0EX, England \\
fmurtagh@acm.org}

\maketitle 

\begin{abstract}
We study two aspects of information semantics: 
(i) the collection of all relationships,
(ii) tracking and spotting anomaly and change.
The first is implemented by endowing all
relevant information spaces with a Euclidean metric in a common 
projected space.  The second is modelled by an induced ultrametric.  
A very general way to achieve a Euclidean embedding of different 
information spaces based on cross-tabulation counts (and from 
other input data formats) is provided by Correspondence Analysis.  From there, 
the induced ultrametric that we are particularly interested in 
takes a sequential -- e.g.\ temporal -- ordering of the data into 
account.  We employ such a perspective to look at narrative, 
``the flow of thought and the flow of language'' (Chafe).  
In application to policy decision making, we
show how we can focus analysis in a small number of dimensions.
\end{abstract}

\begin{figure}
\begin{center}
PHOTO NOT AVAILABLE IN THIS VERSION 
\end{center}
\caption{
{\em Photo shows from left to right: 
Prof.\ John Morrison (Director BCRI), Prof.\ Patrick Fitzpatrick
(Director BCRI), Prof.\ Fionn Murtagh,  Dr.\
James Grannell, Chairman, School of Mathematical Sciences, Prof.
Eugene Freuder (Director Cork Constraint Computation Centre).}
The Annual Boole Lecture was established and is sponsored by 
the Boole Centre for Research in Informatics, the Cork Constraint
Computation Centre, the Department of Computer Science, and the 
School of Mathematical Sciences, at
University College Cork.  The series in named in honour of George
Boole, the first professor of Mathematics at UCC, whose seminal
work on logic in the mid-1800s is central to modern digital 
computing.}
\label{fig0}
\end{figure}

\section{Analysis of Narrative}

\subsection{Introduction}

The data mining and data analysis challenges addressed are the 
following. 

\begin{itemize}
\item Great masses of data, textual and otherwise, need to be 
exploited and decisions need to be made.  Correspondence Analysis 
handles multivariate numerical and symbolic data with ease. 
\item Structures and interrelationships evolve in time.
\item We must consider a complex web of relationships.
\item We need to address all these issues from data sets and data flows.   
\end{itemize}

Various aspects of how we respond to these challenges will be discussed 
in this article, complemented by the Appendix.  
We will look at how this works, using the Casablanca film script.
Then we return to the data mining approach used, to propose that various
issues in policy analysis can be addressed by such techniques also.

\subsection{The Changing Nature of Movie and Drama}

McKee \cite{mckee} bears out the great importance of the 
film script: ``50\% of what we understand comes from watching it being 
said.''  And: ``A screenplay waits for the camera. ... Ninety percent 
of all verbal expression has no filmic equivalent.''

An episode of a television series costs \$2--3 million per 
one hour of television, or \pounds 600k--800k for a similar series in the UK. 
Generally screenplays are written speculatively or commissioned, and then 
prototyped by the full production of a pilot episode.  Increasingly, and
especially availed of by the young, television series are delivered via the 
Internet.  

Originating in one medium -- cinema, television, game, online
-- film and drama series are increasingly migrated to another.   
So scriptwriting must take account of digital multimedia platforms.  
This has been referred to in computer networking parlance as ``multiplay''
and in the television media sector as a ``360 degree'' environment.  

Cross-platform delivery motivates interactivity in drama.  So-called
reality TV has a considerable degree of interactivity, as well as being
largely unscripted.  

There is a burgeoning need for us to be in a position to model the
semantics of film script, -- its most revealing structures, patterns
and  layers.
With the drive towards interactivity, we also want to leverage 
this work towards more general scenario analysis.  Potential 
applications are to business strategy and planning; education 
and training; and science, technology and economic development policy.
We will discuss initial work on the application to policy decision
making in section \ref{sect3} below.  

\subsection{Correspondence Analysis as a Semantic Analysis Platform}

For McKee \cite{mckee},  film script 
text is the ``sensory surface of a work of 
art'' and reflects the underlying emotion or perception.  Our data
mining approach models and tracks these underlying aspects in 
the data.  Our approach to textual data mining has a range of novel elements.
  
Firstly, a novelty is our focus on the orientation of narrative 
through Correspondence Analysis \cite{benz,murtagh05}
which maps scenes (and subscenes), and words used, in a near fully 
automated way, into a Euclidean space representing all pairwise 
interrelationships.  Such a space is ideal for visualization.  
Interrelationships between scenes are captured and displayed, as 
well as interrelationships between words, and mutually between 
scenes and words.   

The starting point for analysis 
is frequency of occurrence data, 
typically the ordered scenes crossed by all words used in the script.  

If the totality of inter-relationships is one facet of semantics, 
then another is anomaly or change as modelled by a clustering 
hierarchy.  If, therefore, a scene is quite different from immediately 
previous scenes, then it will be incorporated into the hierarchy at a 
high level.  This novel view of hierarchy will be discussed further 
in section \ref{sectgeotop} below.   

We draw on these two vantage points on semantics -- 
viz. totality of inter-relationships, and using a hierarchy to express change.
  
Among further work that we report on in \cite{murtaghetal08} is 
the following.  
We devise a Monte Carlo approach to test statistical significance of 
the given script's patterns and structures as opposed to randomized 
alternatives (i.e.\ randomized realizations of the scenes).   
Alternatively we examine caesuras and breakpoints in the film script, 
by taking the Euclidean embedding 
further and inducing an ultrametric on the sequence of scenes. 

\subsection{Casablanca Narrative: Illustrative Analysis}

The well known Casablanca movie serves as an example for us.  
Film scripts, such as for Casablanca, are partially structured
texts.  Each scene has metadata and the body of the scene contains
 dialogue and possibly other descriptive data.  
The Casablanca script was half completed when production began in 
1942.  The dialogue for some scenes was written while shooting was in 
progress.  Casablanca was based on an unpublished 1940 screenplay 
\cite{burnett}.  It was scripted by J.J. Epstein, P.G. 
Epstein and H. Koch.  The film was 
directed by M. Curtiz and produced by H.B. Wallis and J.L. Warner.
It was shot by Warner Bros.\ between May and August 1942.

As an illustrative first example we use the following.
A data set was constructed from the 77 successive scenes crossed by 
attributes -- Int[erior], Ext[erior], Day, Night, Rick, Ilsa, Renault, 
Strasser, Laszlo, Other (i.e.\ minor character), and 29 locations.  
Many locations were met with just once; and Rick's Caf\'e was the 
location of 36 scenes.  In scenes based in Rick's Caf\'e we did 
not distinguish between ``Main 
room'', ``Office'', ``Balcony'', etc.  Because of the plethora of 
scenes other than Rick's Caf\'e we assimilate these to just one, ``other than
Rick's Caf\'e'', scene.

\begin{figure*}
\begin{center}
\includegraphics[width=16cm]{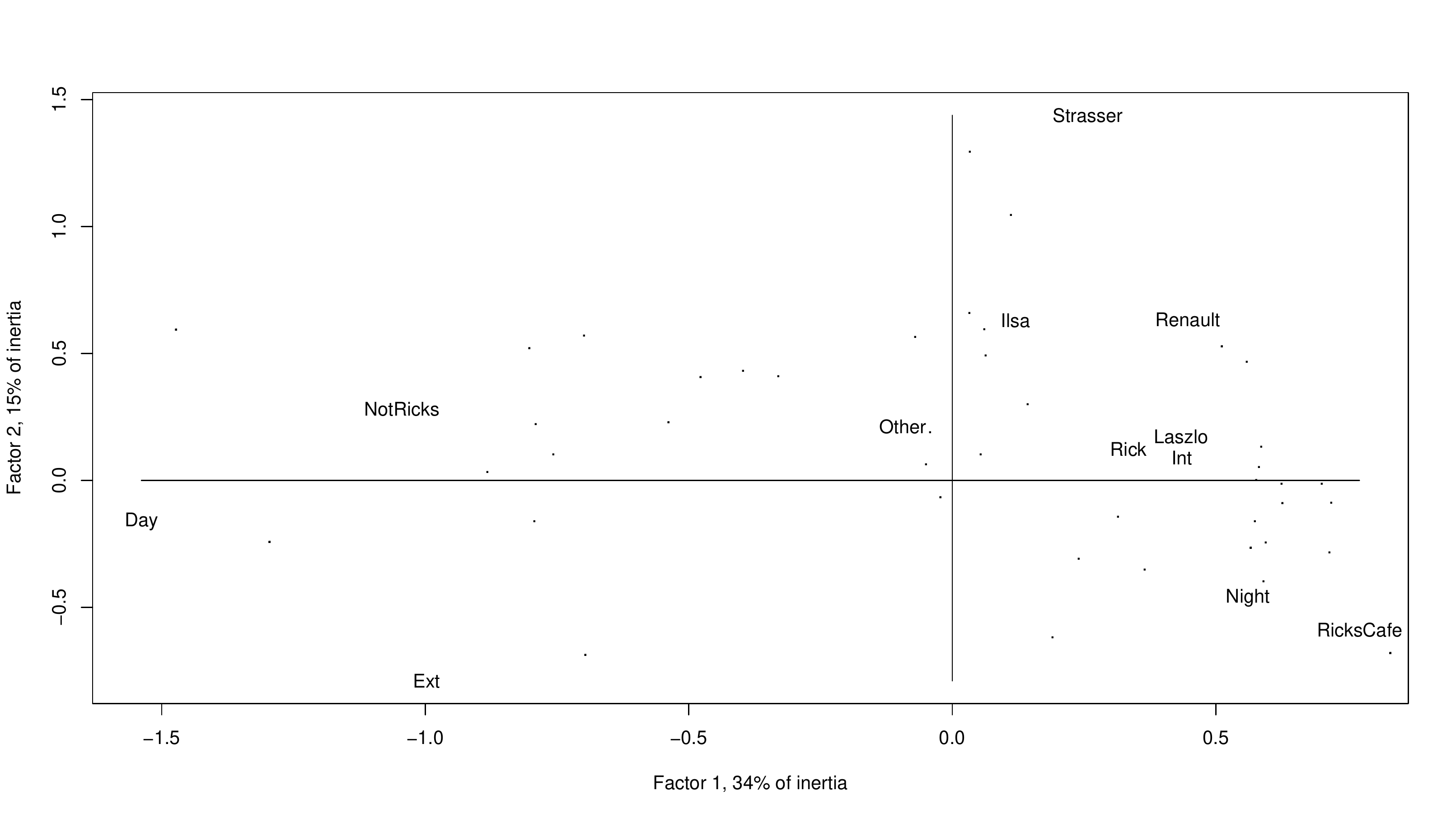}
\end{center}
\caption{Correspondence Analysis of the Casablanca data derived from
the script.  The input data is presences/absences for 77 scenes
crossed by 12 attributes.  The 77 scenes are located at the dots, 
which are not labelled here for clarity.  For a short review of the 
analysis methodology, see Appendix.}
\label{fig1}
\end{figure*}

In Figure \ref{fig1}, 12 attributes are displayed; 77 scenes are 
displayed as dots (to avoid over-crowding of labels).  
Approximately 34\% (for factor 1) + 15\% (for factor 2) = 
49\% of all information, expressed as 
inertia explained, is displayed here. 
We can study interrelationships between characters, other attributes, 
scenes, for instance closeness of Rick's Caf\'e with Night and 
Int (obviously enough).  

\subsection{Modelling Semantics via the Geometry and Topology of 
Information}
\label{sectgeotop}

Some underlying principles are as follows.  
We start with the cross-tabulation data, scenes $\times$ attributes.
Scenes and attributes are embedded in a metric space.
This is how we are probing the {\em geometry of information},
which is a term and viewpoint used by \cite{vanr}.  

\begin{figure}
\begin{center}
\includegraphics[width=7cm]{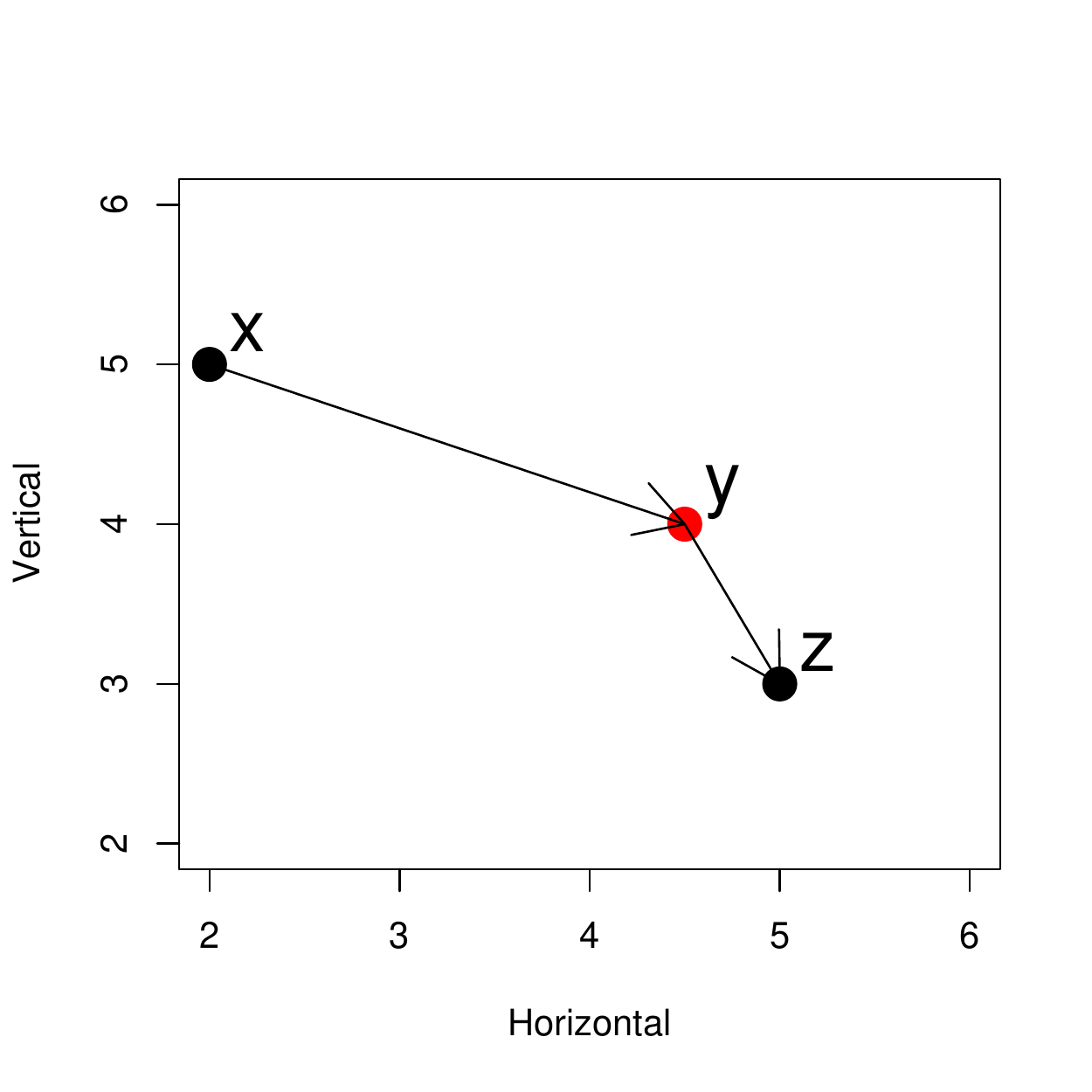}
\end{center}
\caption{The triangular inequality defines a metric: 
every triplet of points satisfies the relationship:
$d(x,z) \leq d(x,y) + d(y,z)$ for distance $d$.}
\label{fig2}
\end{figure}

Underpinning the display in Figure \ref{fig1} is a Euclidean embedding.
The triangular inequality holds for metrics.  An example of a metric is 
the Euclidean distance, exemplified in Figure \ref{fig2}, where each 
and every triplet of points satisfies the relationship:
$d(x,z) \leq d(x,y) + d(y,z)$ for distance $d$.  Two other relationships
also must hold.  These are symmetry and positive definiteness, respectively: 
$d(x,y) = d(y,x)$, and $d(x,y) > 0 $ if $x \neq y$, $d(x,y) = 0 $ if $x = y$.

Further underlying principles used in Figure \ref{fig1} are as follows.
The axes are the principal axes of momentum.
Identical principles are used as in classical mechanics.
The scenes are located as weighted averages of all associated 
attributes; and vice versa. 

\begin{figure}
\begin{center}
\includegraphics[width=8cm]{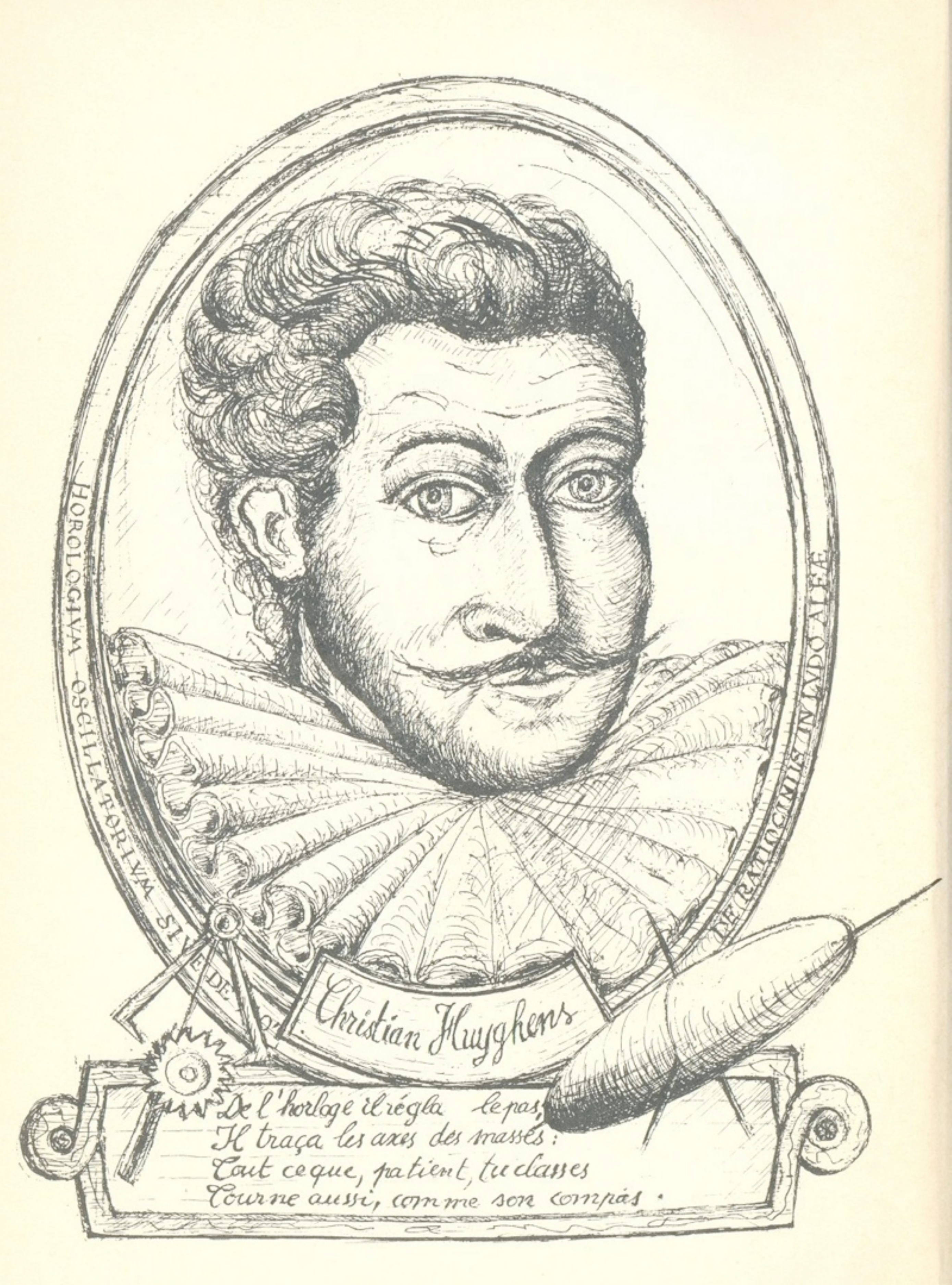}
\end{center}
\caption{Depiction of Christiaan Huyghens, 1629--1695,
 from \cite{benz}.
Towards the bottom on the right there is a depiction of the 
decomposition of the inertia of a hyperellipsoid cloud.}
\label{fig3}
\end{figure}

Huyghens' theorem (cf.\ Figure \ref{fig3}) relates to decomposition of 
inertia of a cloud of points. This is the basis of Correspondence Analysis.

We come now to a different principle: that of the {\em topology of 
information}.   The particular topology used is that of hierarchy.
Euclidean embedding provides a very good starting point to look at 
hierarchical relationships.
An innovation in our work is as follows: the hierarchy takes sequence, e.g.\
timeline, into account.
This captures, in a more easily understood way, the notions of
novelty, anomaly or change.

\begin{figure}
\begin{center}
\includegraphics[width=7cm]{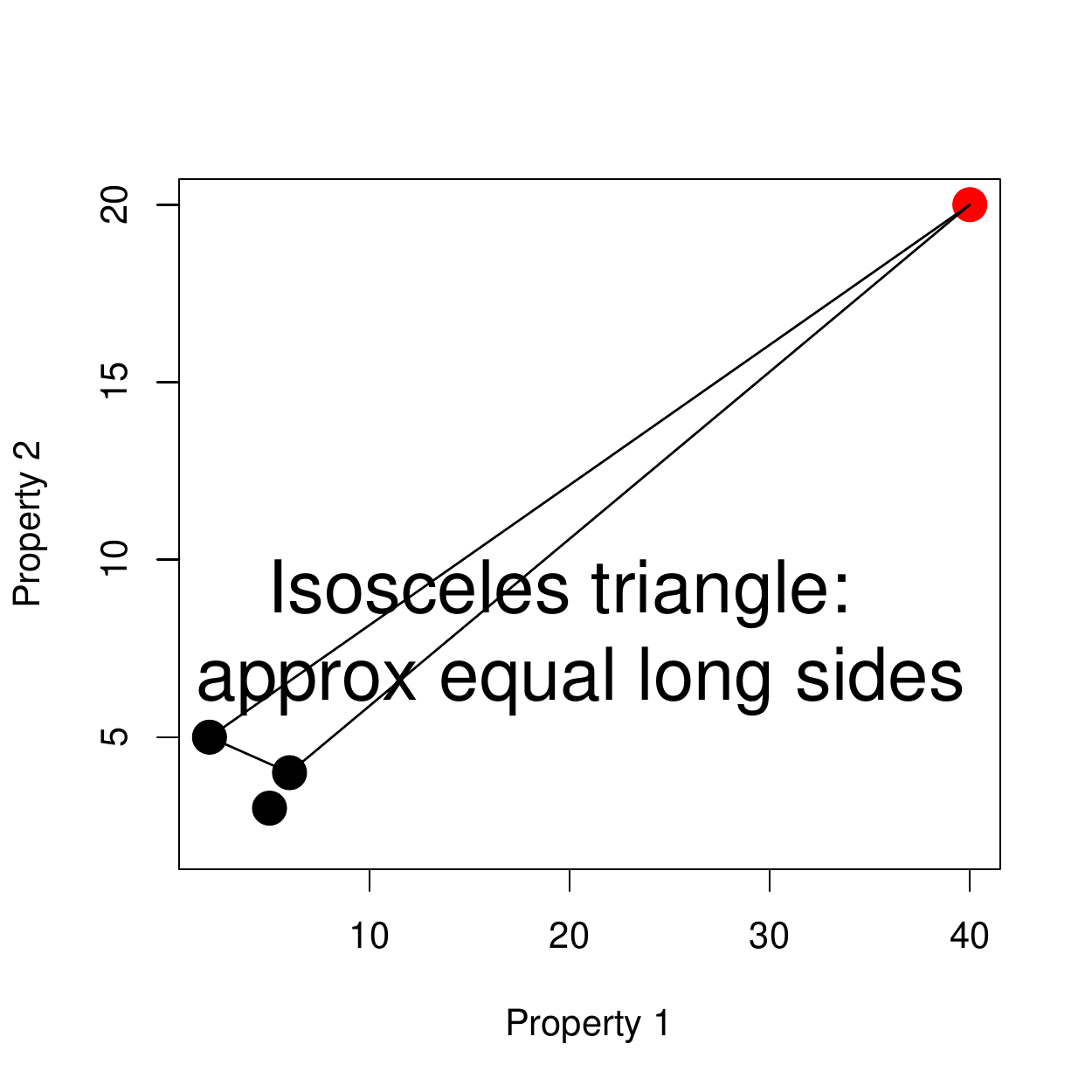}
\end{center}
\caption{The query is on the far right.  While we can easily 
determine the closest target (among the three objects represented
by the dots on the left), is the closest really that much different
from the alternatives?}
\label{fig4}
\end{figure}

Let us take an informal case study to see how this works.  
Consider the situation of seeking documents based on titles.  If the target
population has at least one document that is close to the query, 
then this is (let us assume) clearcut.  However if all documents in the 
target population are very unlike the query, does it make any sense to 
choose the closest?  Whatever the answer here we are focusing on the 
inherent ambiguity, which we will note or record in an appropriate way.  
Figure \ref{fig4} illustrates this situation, where 
the query is the point to the right.   

By using approximate similarity this situation can be 
modelled as an isosceles triangle with small base, as illustrated in
Figure \ref{fig4}.  
An ultrametric space has properties that are very unlike a metric
space, and one such property is that the only triangles allowed are either
(i) equilateral, or (ii) isosceles with small base.  So Figure \ref{fig4}
can be taken as representing a case of ultrametricity.  What this means
is that the query can be viewed as having a particular sort of dominance
or hierarchical relationship vis-\`a-vis any pair of target documents.
Hence any triplet of points here, one of which is the query (defining the 
apex of the isosceles, with small base, triangle), defines local hierarchical
or ultrametric structure.  (See \cite{murtagh04} for case studies.) 

It is clear from Figure \ref{fig4} that we should use approximate
equality of the long sides of the triangle.  The further away the query is 
from the other data then the better is this approximation \cite{murtagh04}.

What sort of explanation does this provide for our conundrum?  It means that 
the query is a novel, or anomalous, or unusual ``document''.   It is up to 
us to decide how to treat such new, innovative cases.  It raises though the 
interesting perspective that here we have a way to model and
subsequently handle the semantics of anomaly or innocuousness.  

\begin{figure}
\begin{center}
\includegraphics[width=8cm]{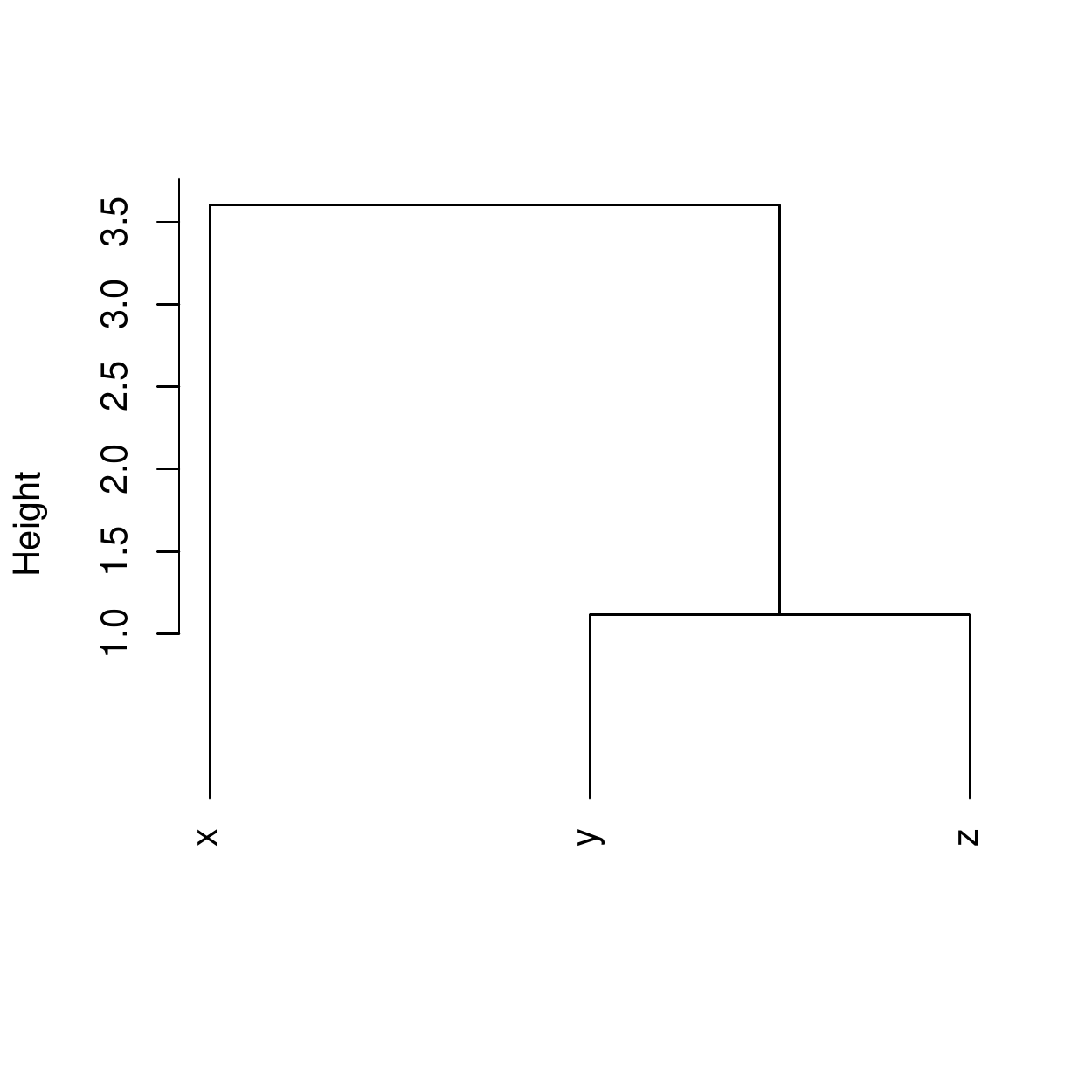}
\end{center}
\caption{The strong triangular inequality defines an ultrametric: 
every triplet of points satisfies the relationship:
$d(x,z) \leq \mbox{max} \{ d(x,y), d(y,z) \}$ for distance $d$.
Cf.\ by reading off the hierarchy, how this is verified for all $x, y, 
z$: $d(x,z) = 3.5; d(x,y) = 3.5; d(y,z) = 1.0$.   
In addition the symmetry and positive definiteness conditions 
hold for any pair of points.}
\label{fig5}
\end{figure}

The strong triangular inequality, or ultrametric inequality, 
holds for tree distances: see Figure \ref{fig5}.  
The closest common ancestor distance is such an ultrametric.

\subsection{Casablanca Narrative: Illustrative Analysis Continued}

\begin{figure*}
\begin{center}
\includegraphics[width=18cm,angle=270]{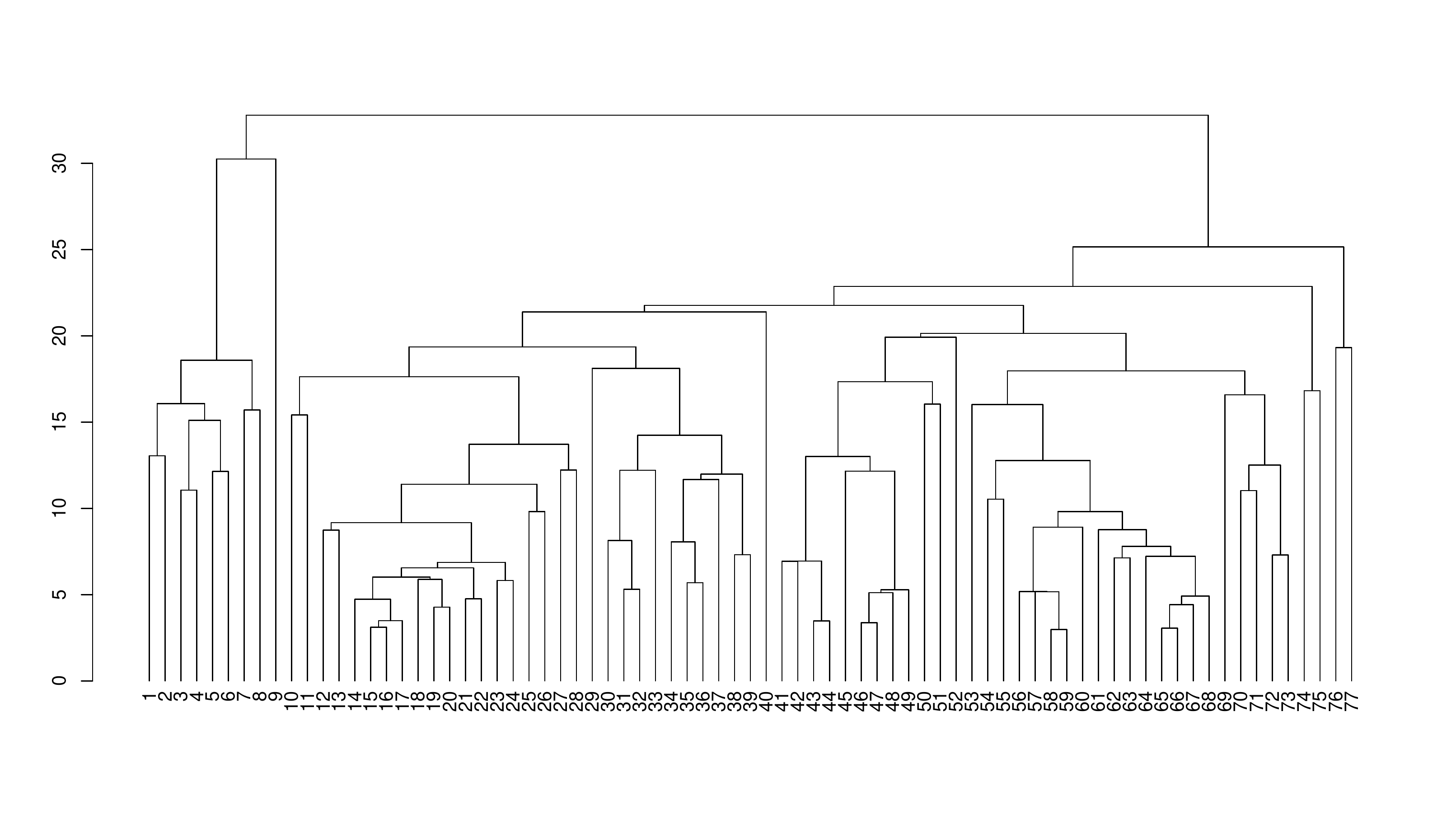}
\end{center}
\caption{77 scenes clustered.  These scenes are in sequence: a 
sequence-constrained agglomerative criterion is used for this.  The 
agglomerative criterion itself is a complete link one.  See \cite{murtagh85}
for properties of this algorithm.}
\label{fig6}
\end{figure*}

Figure \ref{fig6} uses a sequence-constrained complete link agglomerative
algorithm.  It shows up scenes 9 to 10, and progressing from 39, to 40 and 41,
as major changes.  The sequence constrained algorithm, 
i.e.\ agglomerations are 
permitted between adjacent segments of scenes only, is described in an 
Appendix to this article, and in greater detail in \cite{murtagh05}.
The agglomerative criterion used, that is subject to this sequence constraint, 
is a complete link one.  

\subsection{Our Platform for Analysis of Semantics}

Correspondence analysis supports the following:

\begin{itemize}
\item analysis of multivariate, mixed numerical/symbolic data
\item web of interrelationships
\item evolution of relationships over time
\end{itemize}

Correspondence Analysis is in practice {\em a tale of three metrics}
\cite{murtagh05}.  The analysis is based on embedding 
a cloud of points from a space 
governed by one metric into another.  Furthermore the cloud offers 
vantage points of both observables and their characterizations, so 
-- in the case of film script -- for 
any one of the metrics we can effortlessly pass between the space of 
filmscript scenes and attribute set.  The three metrics are as follows. 

\begin{itemize}
\item Chi squared, $\chi^2$, metric -- 
appropriate for profiles of frequencies of occurrence.
\item Euclidean metric, for visualization, and for static context.
\item 
Ultrametric, for hierarchic relations and, as we use it in this work,
for dynamic context.
\end{itemize}

In the analysis of semantics, we distinguish two separate aspects.

\begin{enumerate}
\item Context -- the collection of all interrelationships. 
\begin{itemize}
\item The Euclidean distance makes a lot of sense when the 
population is homogeneous.
\item All interrelationships together provide context, 
relativities -- and hence meaning.  
\end{itemize}

\item Hierarchy tracks anomaly.
\begin{itemize}
\item Ultrametric distance makes a lot of sense when the 
observables are heterogeneous, discontinuous.
\item The latter is especially useful for determining: anomalous, 
atypical, innovative cases.
\end{itemize}
\end{enumerate}

\section{Deeper Look at Semantics of Casablanca}

\subsection{Text Data Mining}

The Casablanca script has 77 successive scenes.
In total there are 6710 words in these scenes.  We define words as 
consisting of at least two letters.  Punctuation is first removed.  
All upper case is set to lower case.  
We use from now on all words.
We analyze frequencies of occurrence of words in scenes, so the input 
is a matrix crossing scenes by words.

\subsection{Analysis of a Pivotal Scene, Scene 43}

As a basis for a deeper look at Casablanca
we have taken comprehensive but qualitative discussion by 
McKee \cite{mckee} and sought quantitative and algorithmic implementation.

Casablanca is based
on a range of miniplots.  For McKee its composition is ``virtually perfect''.

Following McKee \cite{mckee} we will carry out an analysis of Casablanca's 
``Mid-Act Climax'', Scene 43, subdivided into 11 ``beats''.
McKee divides this scene, relating to Ilsa and Rick seeking black market 
exit visas, into 11 ``beats''.

\begin{enumerate}
\item Beat 1 is Rick finding Ilsa in the market.
\item Beats 2, 3, 4 are rejections of him by Ilsa.
\item Beats 5, 6 express rapprochement by both.
\item Beat 7 is guilt-tripping by each in turn.
\item Beat 8 is a jump in content: Ilsa says she will leave Casablanca soon.
\item In beat 9, Rick calls her a coward, and Ilsa calls him a fool.
\item In beat 10, Rick propositions her. 
\item In beat 11, the climax, all goes to rack and ruin: Ilsa says she was
married to Laszlo all along.  Rick is stunned.
\end{enumerate}

\begin{figure*}
\begin{center}
\includegraphics[width=15cm]{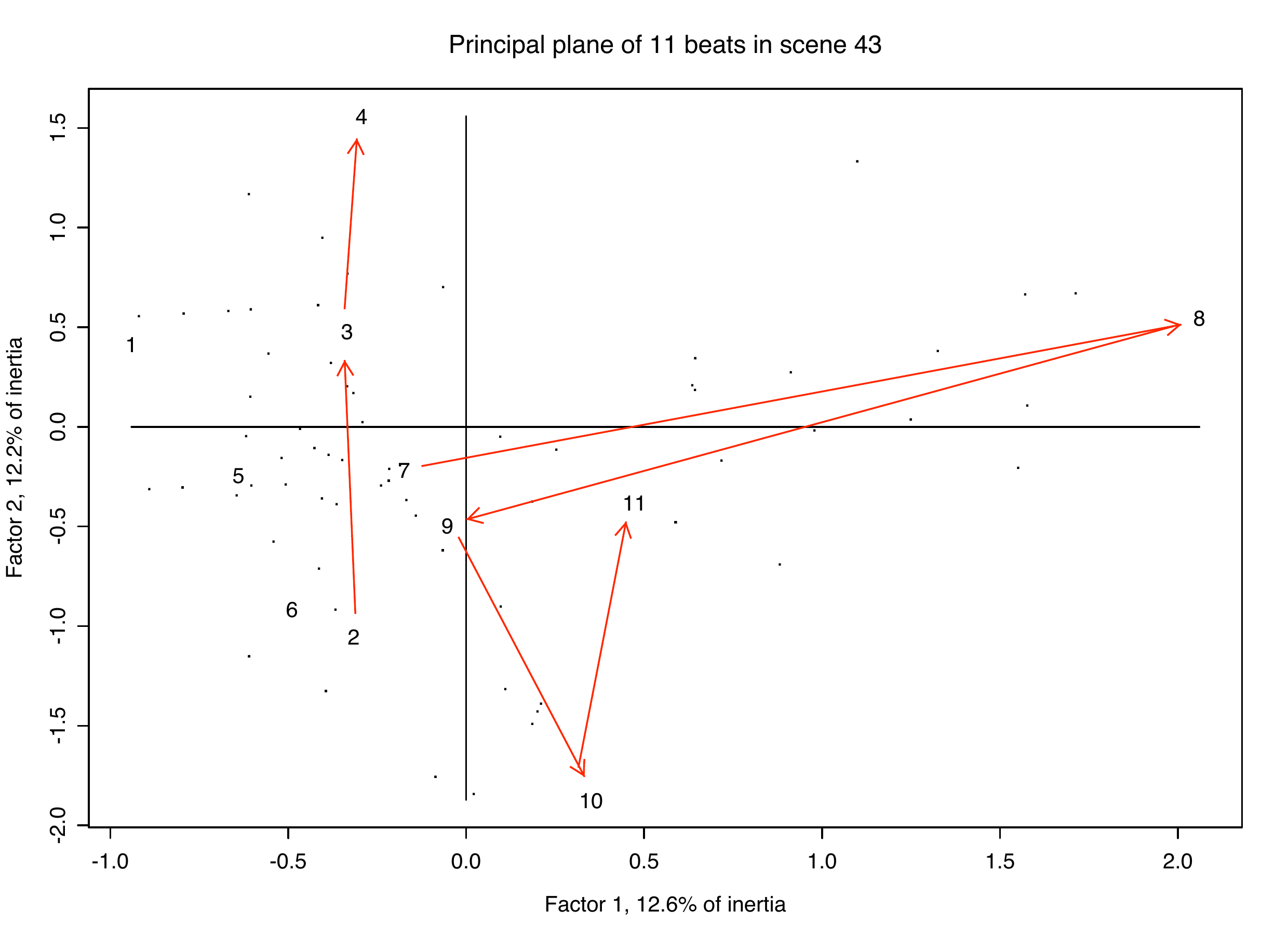}
\end{center}
\caption{Correspondence Analysis principal plane -- best Euclidean
embedding in two dimensions --  of scene 43.  This scene is a central 
and indeed a pivotal one in the movie Casablanca.  It consists of 
eleven sub-scenes, which McKee terms ``beats''.  We discuss in the text 
the evolution over sub-scenes 2, 3 and 4; and again over sub-scenes 
7, 8, 9, 10, and 11.}
\label{fig7}
\end{figure*}

Figure \ref{fig7} shows the evolution from beat to beat rather well.
210 words are used in these 11 ``beats'' or subscenes.  Beat 8 is 
a dramatic development.  Moving upwards on the ordinate (factor 2) 
indicates distance between Rick and Ilsa.   Moving downwards indicates
rapprochement.  

In the full-dimensional space we can check some other of McKee's guidelines.
Lengths of beat get shorter leading up to climax: word counts of final five 
beats in scene 43 are: 50 -- 44 -- 38 -- 30 --- 46.
A style analysis of scene 43 based on McKee can be 
Monte Carlo tested against 999 uniformly randomized sets of the beats.
In the great majority of cases (against 83\% and more of the 
randomized alternatives) we find the style in scene 43 to be characterized by:
small variability of movement from one beat to the next;
greater tempo of beats; and 
high mean rhythm.  

The planar representation in Figure \ref{fig7} 
accounts for approximately 12.6\% + 12.2\% = 24.8\% of the inertia, and 
hence the total information.  
We will look at the evolution of this scene, scene 43, using 
hierarchical clustering of the full-dimensional data -- but based on 
the relative orientations, or correlations with factors.  This is because
of what we have found in Figure \ref{fig7}, viz.\ {\em change} of direction is 
most important. 

\begin{figure*}
\begin{center}
\includegraphics[width=10cm]{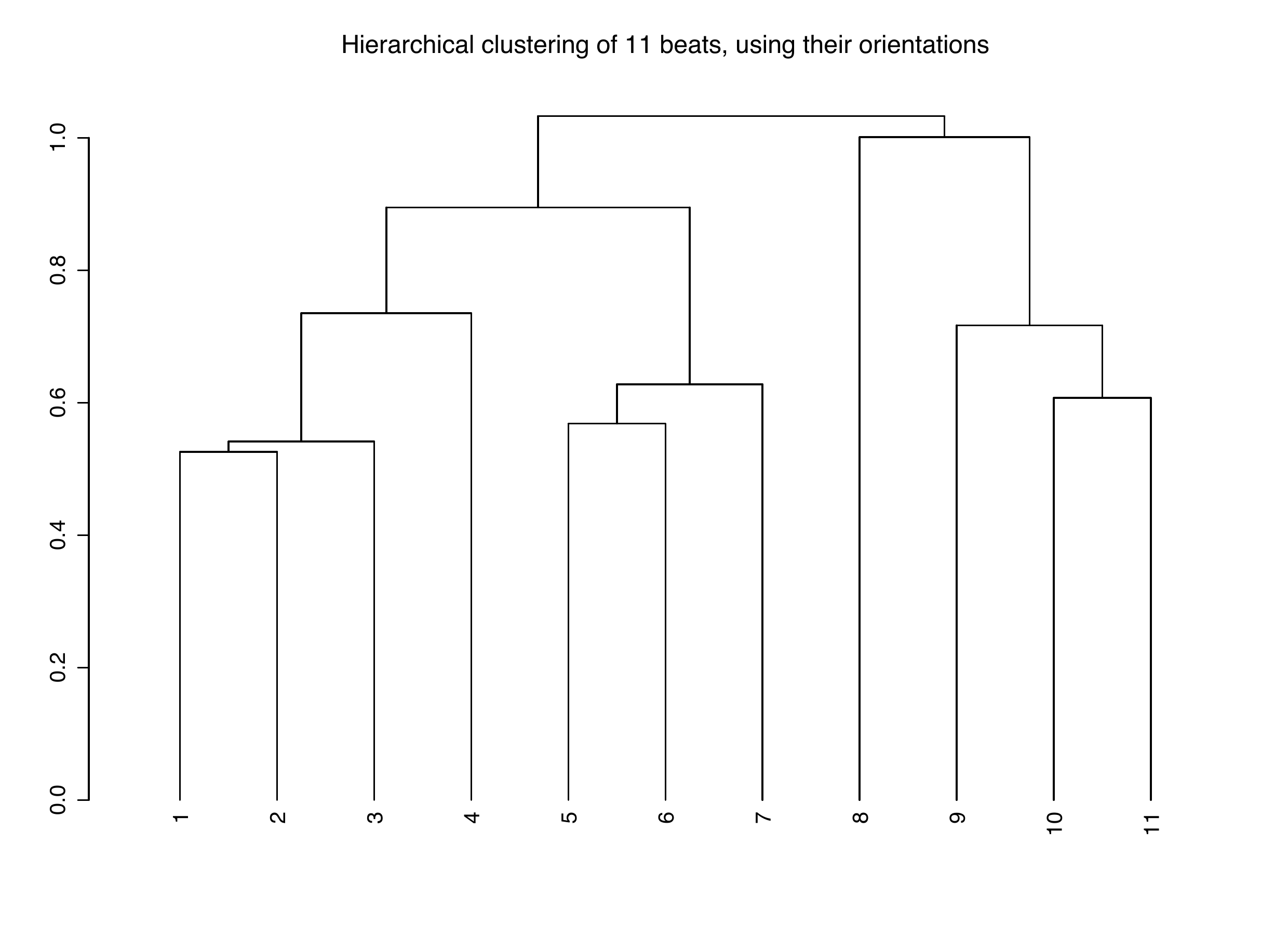}
\end{center}
\caption{Hierarchical clustering of sequence of beats in scene 43 of 
Casablanca.  Again, a sequence-constrained complete link agglomerative
clustering algorithm is used.  The input data is based on 
the full dimensionality
Euclidean embedding provided by the Correspondence Analysis.  The 
relative orientations (defined by correlations with the factors) are used 
as input data.  }
\label{fig8}
\end{figure*}

Figure \ref{fig8} shows the hierarchical clustering, 
based on the sequence of beats.  Input data are of
full dimensionality so there is no approximation involved.  
Note the caesura in moving from beat 7 to 8, 
and back to 9.   There is less of a caesura 
in moving from 4 to 5 but it is still quite pronounced.

Further discussion of these results can be found in \cite{murtaghetal08}.

\section{Application of Narrative Analysis to Science and Engineering Policy}
\label{sect3}

Our way of analyzing semantics is sketched out as follows: 

\begin{itemize}
\item 
We discern story semantics arising out of the orientation of narrative.
\item This is based on the web of interrelationships.
\item We examine caesuras and breakpoints in the flow of narrative.
\end{itemize}

Let us look at the implications of this for data mining with 
decision policy support in view.  

Consider a fairly typical funded project, and its phases up to and beyond
the funding decision.  Different research funding agencies differ in their
procedures.  But a narrative can always be strung together.  
All stages of the proposal and successful project life cycle, 
including external evaluation and internal decision making, are highly
document -- and as a consequence narrative -- based.  It is 
precisely the analysis of such a narrative that is our longer term goal.

As a first step towards much fuller analysis of these many and varied
narratives involved in funded projects, let us look at the very 
general role of narrative in 
national research development.   We will look at: 

\begin{itemize}
\item Overall view -- overall synthesis of information
\item Orientation of strands of development
\item Their tempo, rhythm
\end{itemize}

Through such an analysis of narrative, among the issues to be addressed are:

\begin{itemize}
\item Strategy and its implementation in 
terms of themes and subthemes represented 
\item Thematic focus and coverage 
\item Organisational clustering
\item Evaluation of outputs in a global context
\item All the above over time
\end{itemize}

Our aim is to understand the ``big picture''.  It is not to replace 
the varied measures of success that are applied, such as publications,
patents, licences, numbers of PhDs completed, company start-ups, and so on.
It is instead to appreciate the broader configuration and 
orientation, and to determine the most salient aspects underlying the 
data.

\subsection{Assessing Coverage and Completeness}
\label{sect31}

SFI Centres for Science, Engineering and Technology (CSETs) are 
campus-industry partnerships typically funded at up to \EUR 20
million over 5 years.  Strategic Research Clusters (SRCs) are also 
research consortia, with industrial partners and over 5 years are
typically funded at up to \EUR 7.5 million.  

We cross-tabulated 8 CSETs and 12 SRCs by a range of terms derived from 
title and summary information; together with budget, numbers of PIs
(Principal Investigators), Co-Is (Co-Investigators), and PhDs.

We can display any or all of this information on a common map, for 
visual convenience a planar display, using Correspondence Analysis.

In mapping SFI CSETs and SRCs, we will now show how 
Correspondence Analysis is based on the upper (near root) 
part of an ontology or concept hierarchy.  This we view as 
{\em information focusing}.
Correspondence Analysis provides simultaneous representation of 
observations and attributes.
Retrospectively, we can project other observations or attributes 
into the factor space: these are supplementary observations or attributes.
A 2-dimensional or planar view is likely to be a gross approximation of 
the full cloud of observations or of attributes.
We may accept such an approximation as rewarding and informative.  
Another way to address this same issue is as follows. 
We define a small number of aggregates of either observations or 
attributes, and carry out the analysis on them.  We then project the 
full set of observations and attributes into the factor space.
For mapping of SFI CSETs and SRCs a simple algebra of themes as set out in 
the next paragraph achieves this goal.  The upshot is that the 2-dimensional
or planar view is a better fit to the full cloud of observations or of
attributes.  

From CSET or SRC characterization as: Physical Systems (Phys), Logical 
Systems (Log), Body/Individual, Health/Collective, and Data \& Information 
(Data), the following thematic areas were defined.

\begin{enumerate}
\item eSciences = Logical Systems, Data \& Information
\item Biosciences = Body/Individual, Health/Collective
\item Medical = Body/Individual, Health/Collective, Physical Systems
\item ICT = Physical Systems, Logical Systems, Data \& Information 
\item eMedical = Body/Individual, Health/Collective, Logical Systems
\item eBiosciences = Body/Individual, Health/Collective, Data \& Information 
\end{enumerate}

This categorization scheme can be viewed as the upper level of a 
concept hierarchy.  It can be contrasted with the somewhat more 
detailed scheme that we used for analysis of articles in the Computer Journal, 
\cite{murtagh08}.  

CSETs labelled in the Figures are: 
APC, Alimentary Pharmabiotic Centre; BDI, Biomedical Diagnostics Institute; 
CRANN, Centre for Research on 
Adaptive Nanostructures and Nanodevices; CTVR, Centre for Telecommunications 
Value-Chain Research; DERI, Digital Enterprise Research Institute; 
LERO, Irish Software Engineering Research Centre; NGL, Centre for 
Next Generation Localization; and REMEDI, Regenerative Medicine Institute.

\begin{figure}
\begin{center}
\includegraphics[width=8cm]{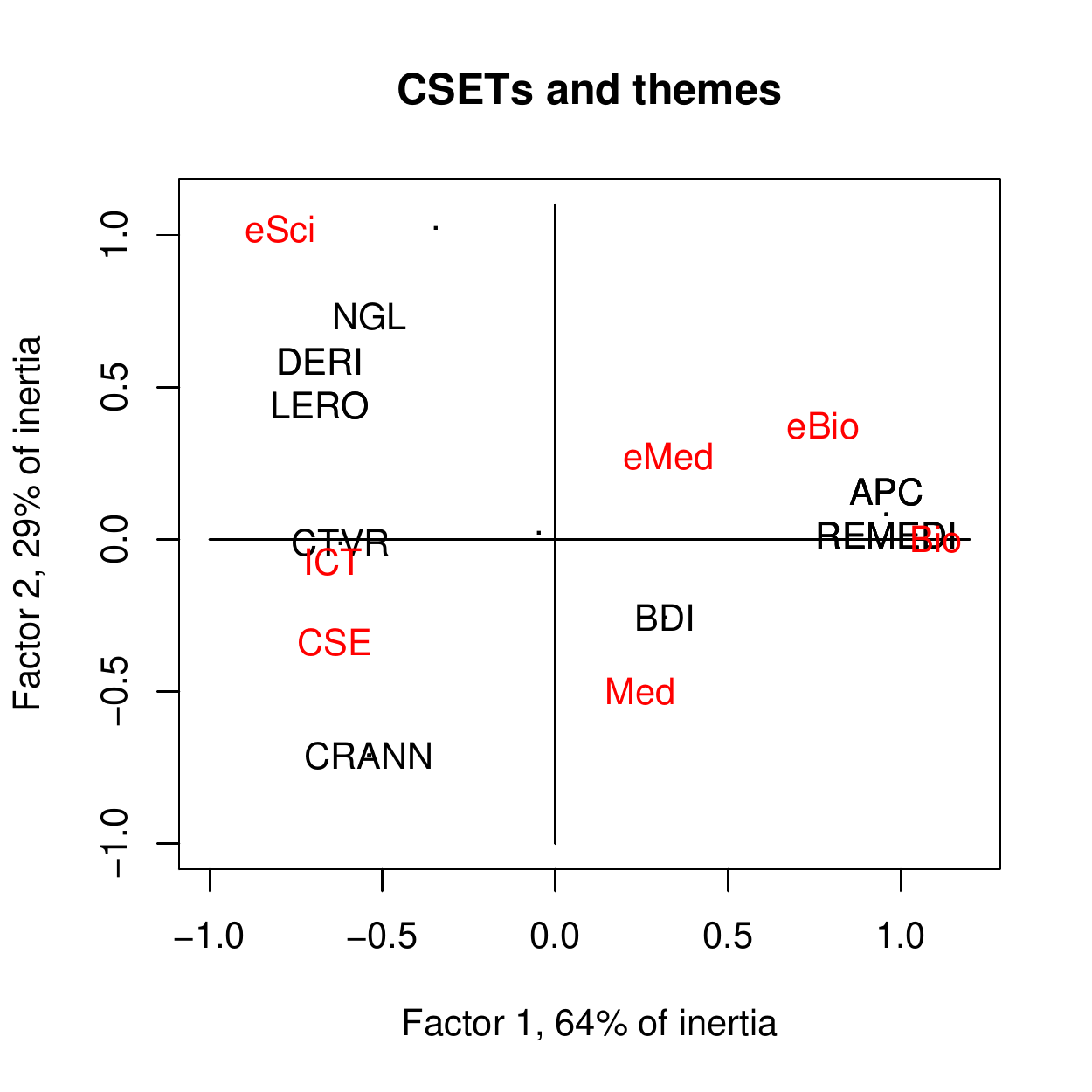}
\end{center}
\caption{CSETs, labelled, with themes located on a planar display, 
which is nearly complete in terms of information content.}
\label{fig9}
\end{figure}

\begin{figure}
\begin{center}
\includegraphics[width=8cm]{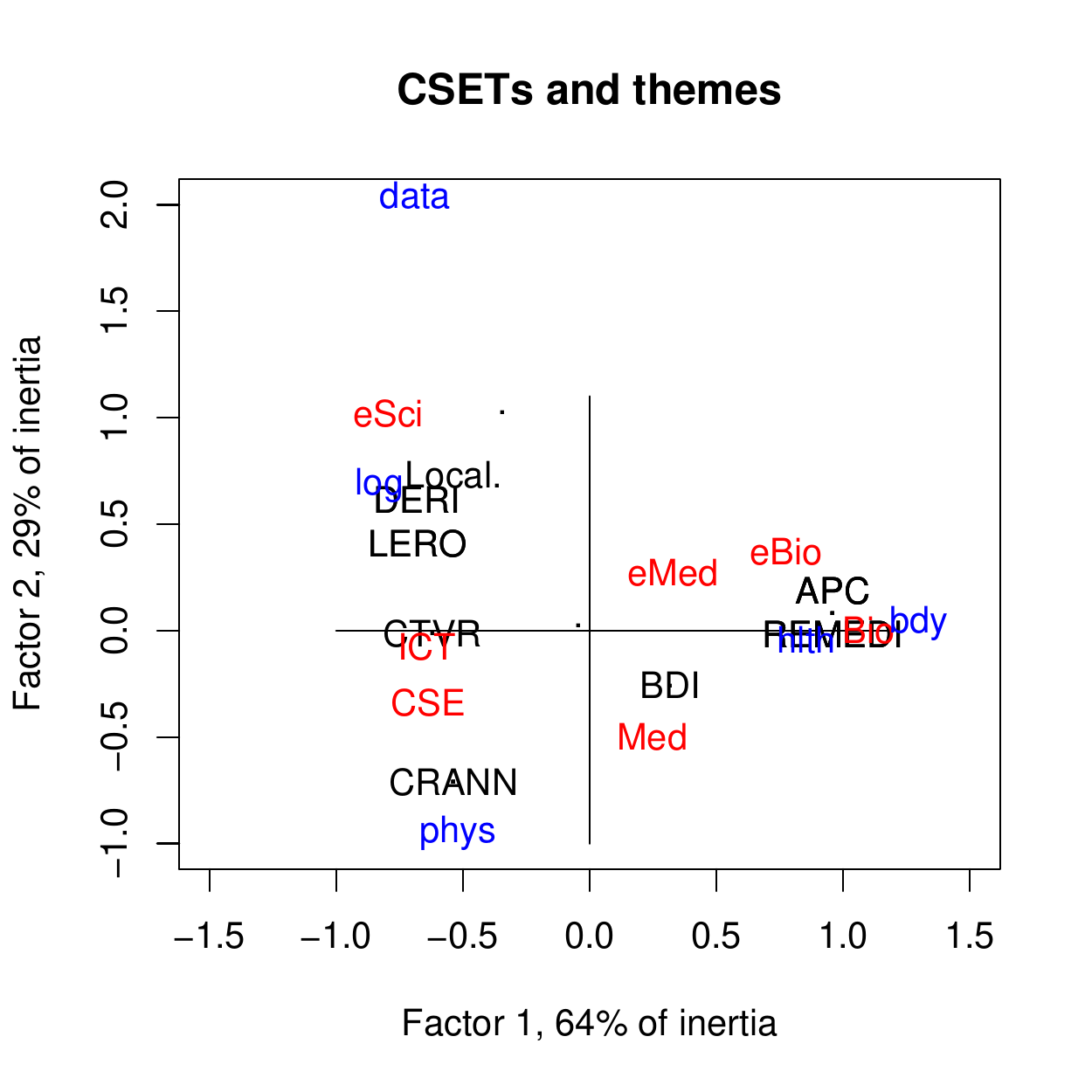}
\end{center}
\caption{As Figure \ref{fig9} but with sub-themes projected into the
display.  Note that, through use of supplementary elements, the axes
and scales are identical to Figures \ref{fig9}, \ref{fig11}, and \ref{fig12}.
Axes and scales are just displayed differently in this figure so 
that sub-themes appear in our field of view.}
\label{fig10}
\end{figure}

In Figure \ref{fig9} eight CSETs and major themes are shown.
Factor 1 counterposes computer engineering (left) to biosciences (right). 
Factor 2 counterposes software on the positive end
to hardware on the negative end.  This 2-dimensional map encapsulates 
64\% (for factor 1) + 29\% (for factor 2) = 93\% of all information, i.e.\
inertia, in the dual clouds of points.  
CSETs are positioned relative to the thematic areas used.  In Figure 
\ref{fig10}, sub-themes are additionally projected into the display.  
This is done by taking the sub-themes as 
{\em supplementary elements} following 
the analysis as such: see Appendix for a short introduction to this.  
From Figure \ref{fig10} we might wish to label additionally factor 2 
as a polarity of data and physics, associated with the extremes of
software and hardware.  

\begin{figure}
\begin{center}
\includegraphics[width=8cm]{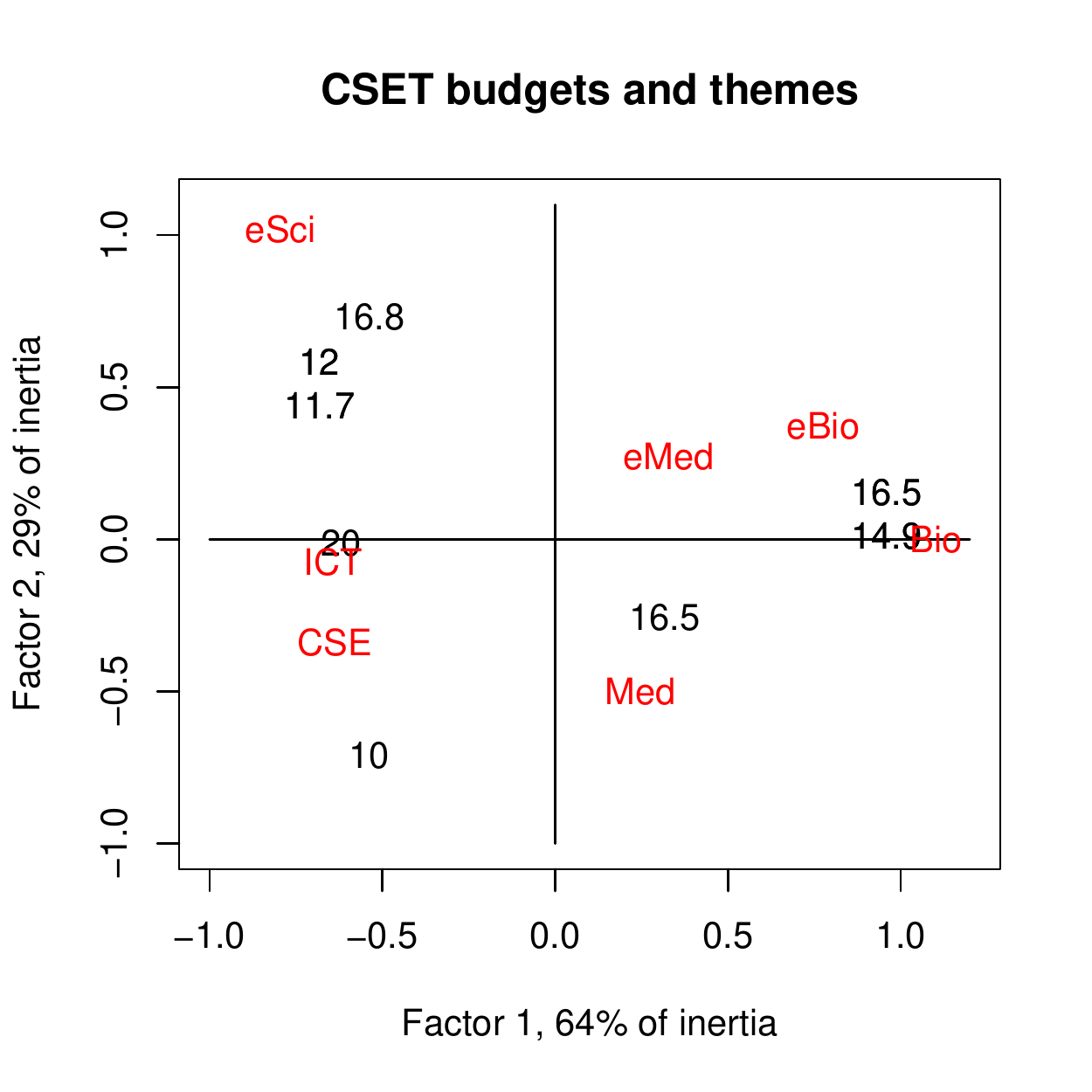}
\end{center}
\caption{Similar to Figure \ref{fig9}, but here we show CSET budgets.}
\label{fig11}
\end{figure}

In Figure \ref{fig11} CSET budgets are shown, as \EUR \ million over 5 years, 
and themes are also displayed.
In this way we use the map to show characteristics 
of the CSETs, in this case budgets.

\begin{figure}
\begin{center}
\includegraphics[width=8cm]{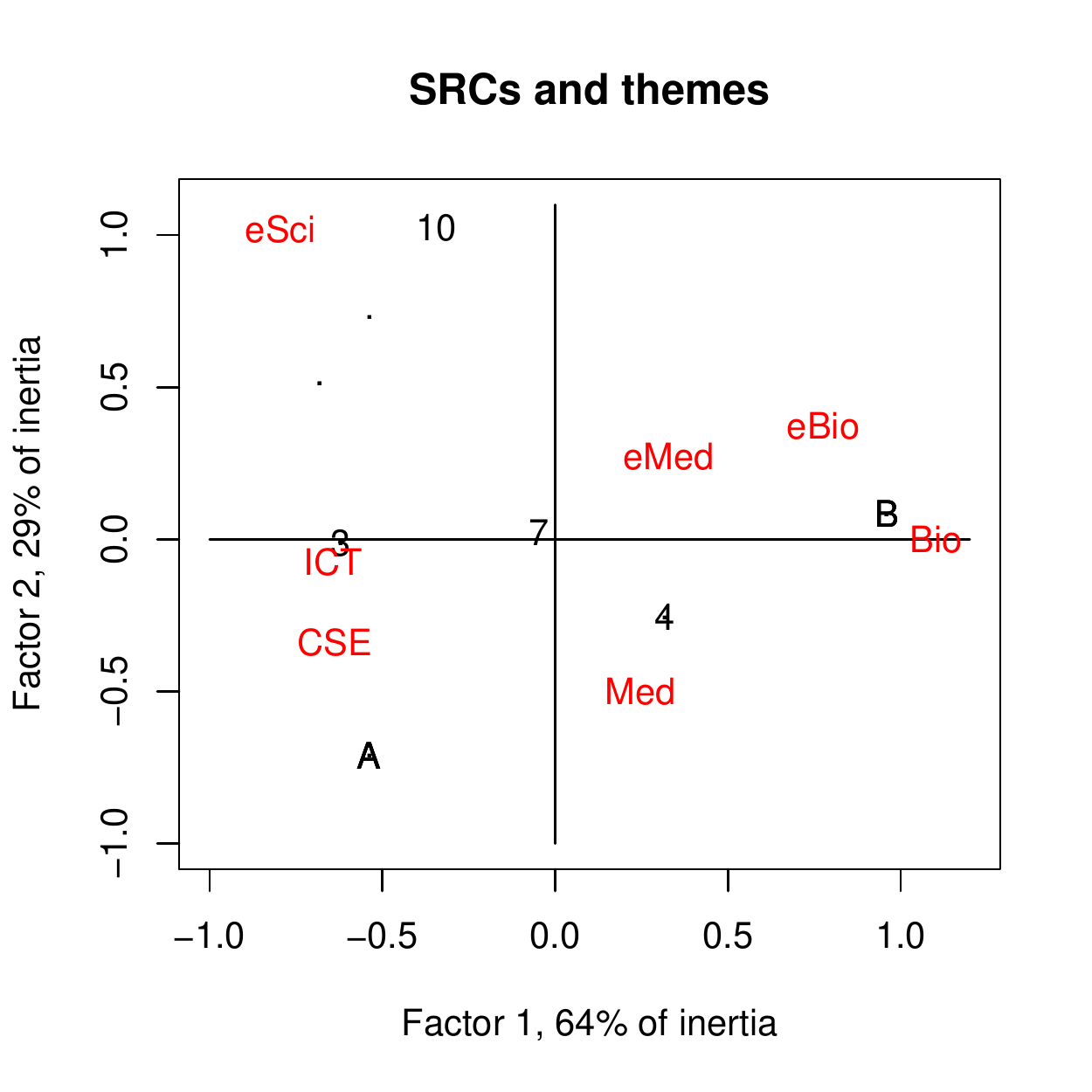}
\end{center}
\caption{Using the same themes, the SRCs are projected.  The properties 
of the planar display are the same as for Figures \ref{fig9}--\ref{fig11}.
SRCs 3, 4, 7, 10 are shown; and overlapping groups of 4 each are at 
A, B.  A represents four bioscience or pharmaceutical SRCs.  B represents four 
materials, biomaterials, or photonics SRCs.  CSETs are displayed as dots
(to avoid overcrowding of labels).}
\label{fig12}
\end{figure}

Figure \ref{fig12} shows 12 SRCs (Strategic Research Clusters)
that  started at the end of 2007.  The planar space into which the 
SRCs are projected is identical to Figures \ref{fig9}--\ref{fig11}. 
This projection is accomplished by supplementary elements (see Appendix).  

\begin{figure}
\begin{center}
\includegraphics[width=8cm]{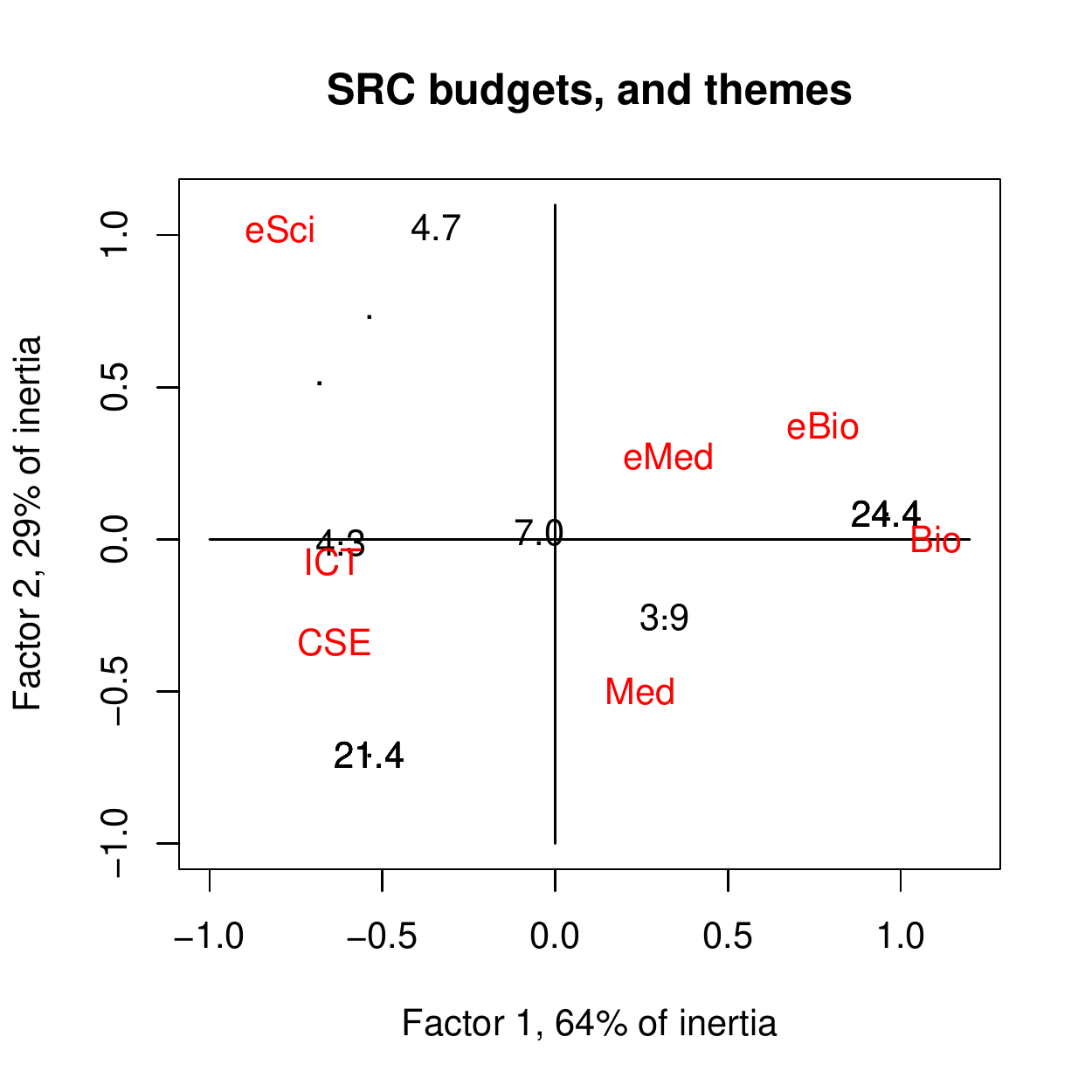}
\end{center}
\caption{As Figure \ref{fig12}, now displaying combined budgets.}
\label{fig13}
\end{figure}

Figure \ref{fig13} shows one property of the SRCs, their budgets 
in \EUR \ million over 5 years.

\subsection{Change Over Time}
\label{sect32}

We take another funding programme, the Research Frontiers Programme,
to show how changes over time can be mapped.  

This programme follows an annual call, and includes 
all fields of science, mathematics and engineering.
There are approximately 750 submissions annually.  
There was a 24\% success rate (168 awards) in 2007, and 
19\% (143 awards) in 2008.
The average award was \EUR 155k in 2007, and \EUR 161k in 2008.
An award runs for three years of funding, and this is moving to four 
years in 2009 to accommodate a 4-year PhD duration.

We will look at the Computer Science panel results only, over 2005, 
2006, 2007 and 2008.

Grants awarded in these years, respectively, were: 14, 11, 15, 17.
The breakdown by institutes concerned was: 
UCD -- 13; TCD -- 10; DCU -- 14; UCC -- 6; UL -- 3; 
DIT -- 3; NUIM -- 3; WIT -- 1.  These institutes are as follows: 
UCD, University College Dublin; DCU, Dublin City University; 
UCC, University College Cork; UL, University of Limerick; NUIM, 
National University of Ireland, Maynooth; DIT, Dublin Institute of 
Technology; and WIT, Waterford Institute of Technology.  

One theme was used to characterize each proposal
from among the following: bioinformatics, 
imaging/video, software, networks, data processing \& information 
retrieval, speech \& language processing, virtual spaces, 
language \& text, information security, and e-learning.
Again this categorization of computer science can be contrasted with 
one derived for articles in recent years in the Computer Journal
\cite{murtagh08}.  

\begin{figure}
\begin{center}
\includegraphics[width=8cm]{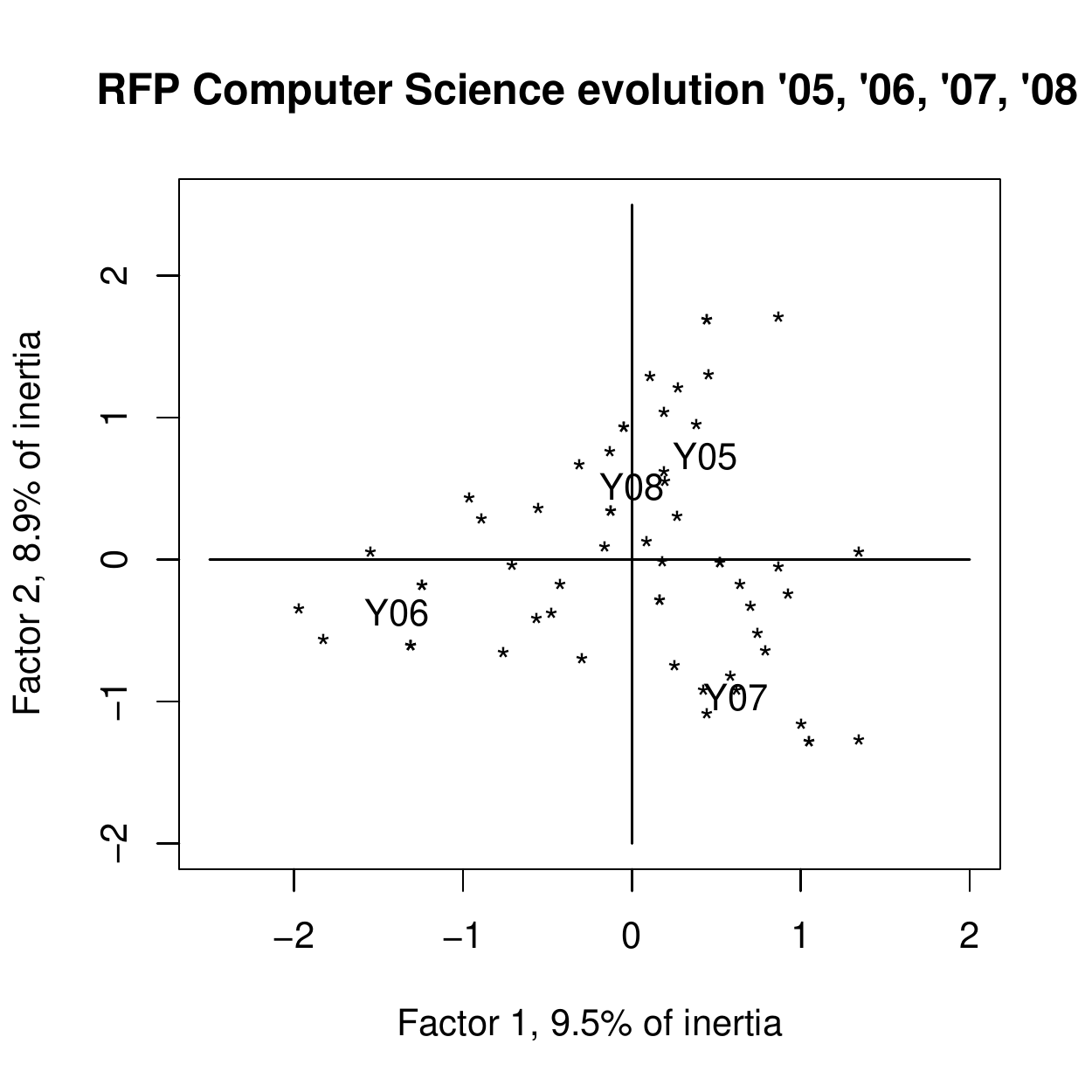}
\end{center}
\caption{Research Frontiers Programme over four years. 
Successful proposals are shown as asterisks.  The years are located 
as the average of successful projects.}
\label{fig14}
\end{figure}

\begin{figure}
\begin{center}
\includegraphics[width=8cm]{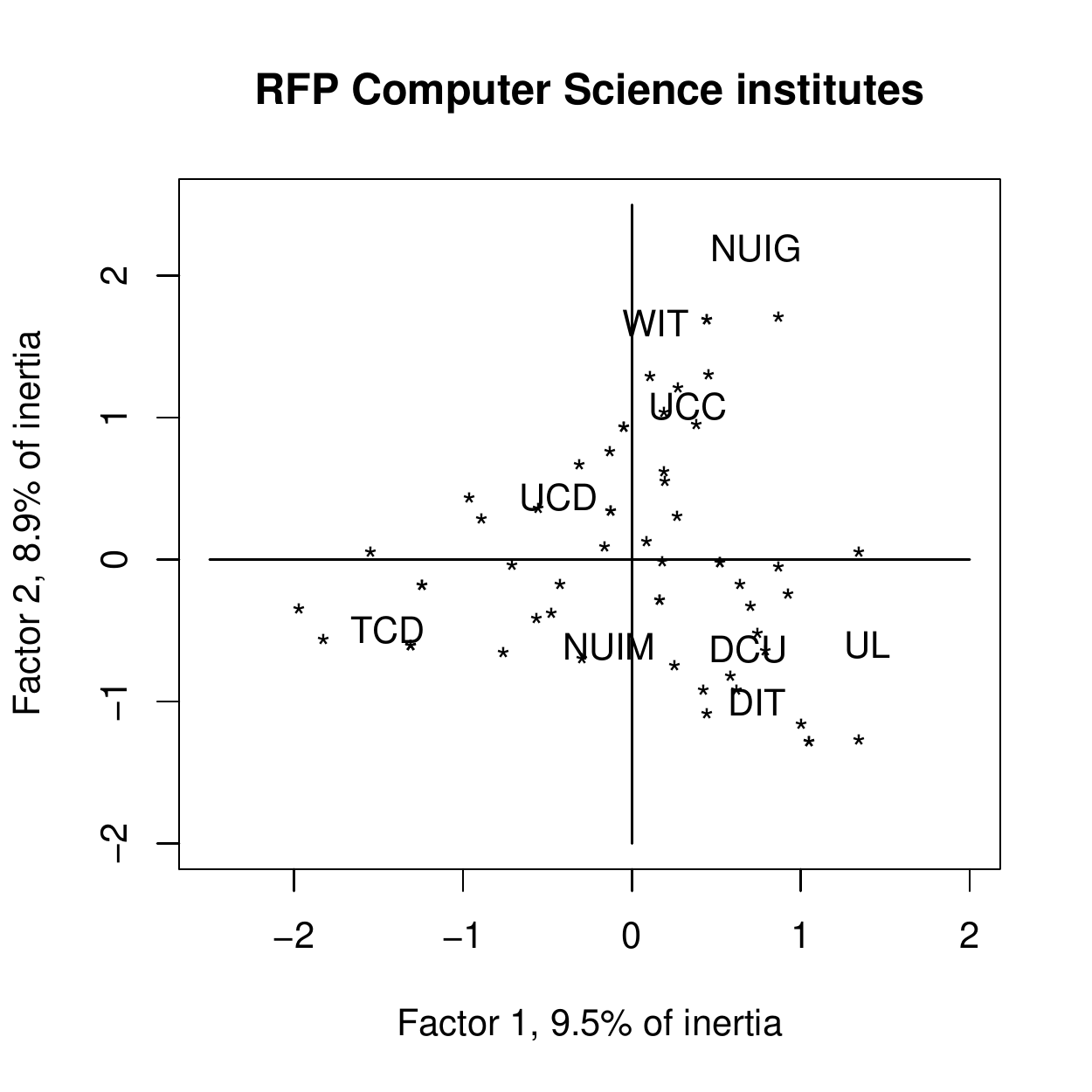}
\end{center}
\caption{As Figure \ref{fig14}, displaying host institutes of the awardees.}
\label{fig15}
\end{figure}

\begin{figure}
\begin{center}
\includegraphics[width=8cm]{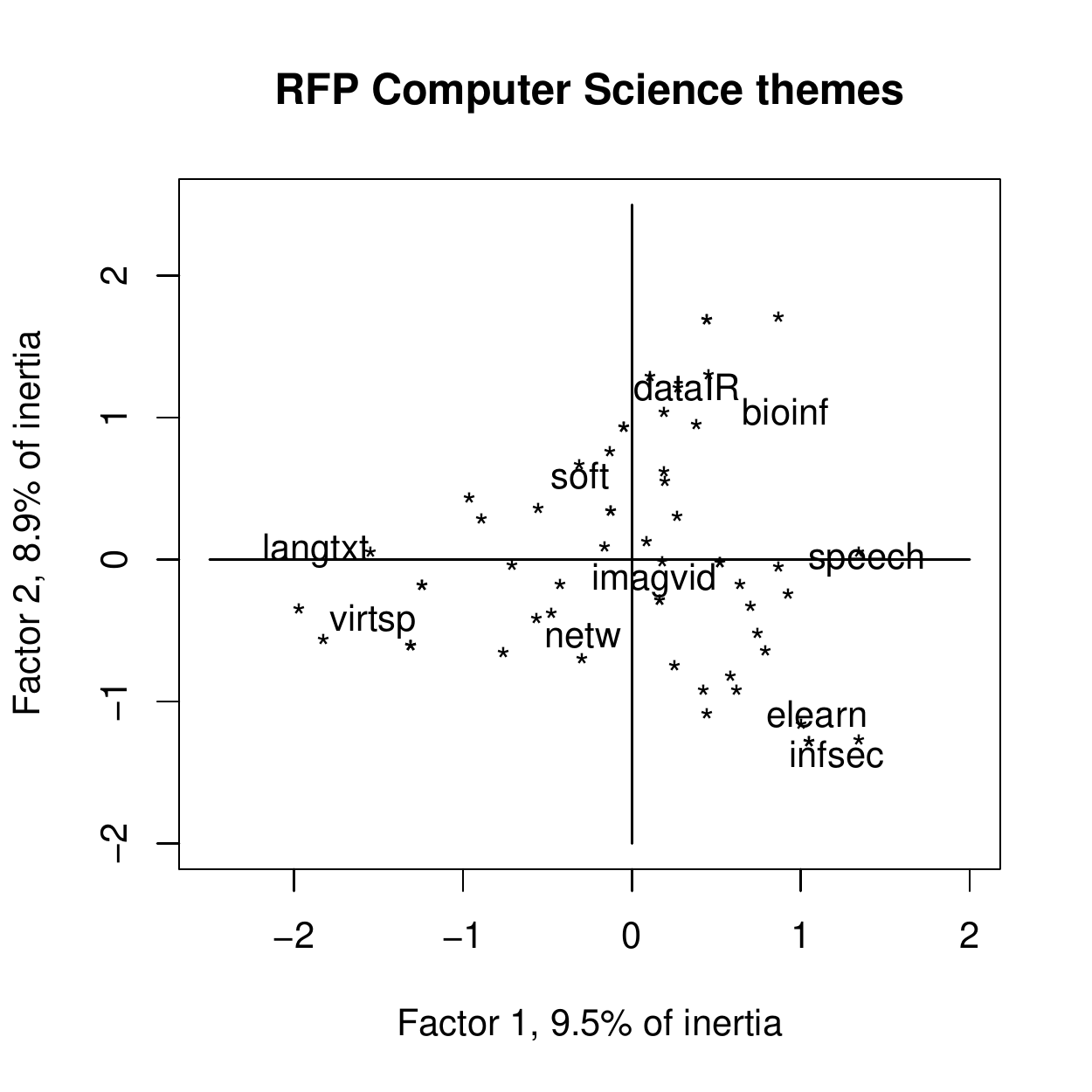}
\end{center}
\caption{As Figures \ref{fig14} and \ref{fig15}, displaying themes.}
\label{fig16}
\end{figure}

Figures \ref{fig14}, \ref{fig15} and \ref{fig16} show different facets of the 
Computer Science outcomes.  By keeping the displays separate, we focus on 
one aspect at a time.  All displays however are based on the same list of 
themes, and so allow mutual comparisons.  Note that the principal 
plane shown accounts for 9.5\% + 8.9\% of the inertia, hence 18.4\% of 
the overall inertia, that in turn expresses information here.  Although 
small, it is the best planar view of the data (arising from the 
$\chi^2$ or chi squared metric, see Appendix,
followed by the Euclidean embedding that the figures 
show).  Ten themes were used, and what the 18.4\% information content 
tells us is that there is importance attached to most if not all of the ten.
We are not prevented though in studying usefully the planar displays.  
That they can be used to display lots of supplementary data is a 
major benefit of their use.  

What our analyses demonstrate is that the categories used are of 
crucial importance.  Indeed, in Figures \ref{fig9}--\ref{fig13}
and then in Figures \ref{fig14}--\ref{fig16}, 
we see how we can ``engineer'' the impact of the 
categories by assimilating their importance to moments of inertia 
of the clouds of associated points.  

\subsection{Conclusion on the Policy Case Studies}

The aims and objectives in our use of the Correspondence Analysis and 
clustering platform is to drive strategy and its implementation in 
policy.  

What we are targeting is to study highly multivariate, evolving data flows. 
This is in terms of the semantics of the data -- principally, complex webs 
of interrelationships and evolution of relationships over time.
This is the {\em narrative of process} that lies behind raw statistics and 
funding decisions.

We have been concerned especially with {\em information focusing} in 
section \ref{sect31}, and this over time in section \ref{sect32}.  

\section{Conclusions}

Data collected on processes or configurations 
can tell us a lot about what is going on.   When backed up with 
probabilistic models, we can test one hypothesis against another.  
This leads to data mining ``in the small''.  Pairwise plots are 
often sufficient because as visualizations they remain close and 
hence faithful to the original data.   Such data mining ``in the small''
is well served by statistics as a discipline.  

Separately from such data mining ``in the small'', there is a 
need for data mining ``in the 
large''.  We need to unravel the broad brush strokes of narrative 
expressing a story.  Indeed as discussed in \cite{chafe}, narrative 
expresses not just a story but human thinking.   A broad brush 
view of science and engineering research, and their linkages with 
the economy, is a similar type of situation.  Pairwise plots are not
always sufficient now, because they tie us down in unnecessary detail.  
We use them since we need the more important patterns, trends, details to be
brought out in order of priority.   Strategy is of the same class 
as narrative.  

For data mining ``in the large'' the choice of data and 
categorization used are crucial.  This we have exemplified well 
in this article.  So our data mining ``in the large'' is tightly 
coupled to the data.  This is how we can make data speak.  
The modelling involved in analysis ``in the 
small'' comes later.

\section*{Appendix: the Correspondence Analysis and Hierarchical Clustering
Platform}

This Appendix introduces important aspects of Correspondence Analysis and 
hierarchical clustering.  Further reading is to be found in 
\cite{benz} and \cite{murtagh05}. 

\subsection*{Analysis Chain}

\begin{enumerate}
\item Starting point: a matrix that cross-tabulates the dependencies,
e.g.\ frequencies of joint occurrence, of an observations crossed by attributes
matrix.  
\item By endowing the cross-tabulation matrix with the $\chi^2$ metric 
on both observation set (rows) and attribute set (columns), we can map 
observations and attributes into the same space, endowed with the Euclidean
metric.  
\item A hierarchical clustering is induced on the Euclidean space, the 
factor space.  
\item Interpretation is through projections of observations, attributes 
or clusters onto factors.  The factors are ordered by decreasing importance.
\end{enumerate}

Various aspects of Correspondence Analysis follow on 
from this, such as Multiple Correspondence Analysis, different ways that
one can encode input data, and mutual description of clusters in terms of 
factors and vice versa.   In the following we use elements of 
the Einstein tensor notation of \cite{benz}.  This often reduces to common
vector notation.  

\subsection*{Correspondence Analysis: 
Mapping $\chi^2$ Distances into Euclidean Distances}

\begin{itemize}
\item The given contingency table (or numbers of occurrence) 
data is denoted $k_{IJ} =
\{ k_{IJ}(i,j) = k(i, j) ; i \in I, j \in J \}$.  

\item $I$ is the set of observation 
indexes, and $J$ is the set of attribute indexes.  We have
$k(i) = \sum_{j \in J} k(i, j)$.  Analogously $k(j)$ is defined,
and $k = \sum_{i \in I, j \in J} k(i,j)$.  

\item Relative frequencies: $f_{IJ} = \{ f_{ij}
= k(i,j)/k ; i \in I, j \in J\} \subset \R_{I \times J}$,
similarly $f_I$ is defined as  $\{f_i = k(i)/k ; i \in I, j \in J\}
\subset \R_I$, and $f_J$ analogously.  
\item The conditional distribution of $f_J$ knowing $i \in I$, also termed
the $j$th {\em profile} with coordinates indexed by the elements of $I$, is:

$$ f^i_J = \{ f^i_j = f_{ij}/f_i = (k_{ij}/k)/(k_i/k) ; f_i > 0 ;
j \in J \}$$ and likewise for $f^j_I$.  

\item What is discussed in terms of information focusing in the text is underpinned
by the {\em principle of distributional equivalence}.  This means that if
two or more profiles are aggregated by simple element-wise summation, then 
the $\chi^2$ distances relating to other profiles are not effected.
\end{itemize}

\subsection*{Input: Cloud of Points Endowed with the Chi Squared Metric}

\begin{itemize}
\item The cloud of points consists of the couples: 
(multidimensional) profile coordinate and (scalar) mass.
We have $ N_J(I) = $
$\{ ( f^i_J, f_i ) ; i  \in I \} \subset \R_J $, and
again similarly for $N_I(J)$.  

\item Included in this expression is the fact
that the cloud of observations, $ N_J(I)$, is a subset of the real 
space of dimensionality $| J |$ where $| . |$ denotes cardinality 
of the attribute set, $J$.  

\item The overall inertia is as follows: 
$$M^2(N_J(I)) = M^2(N_I(J)) = \| f_{IJ} - f_I f_J \|^2_{f_I f_J} $$
$$= \sum_{i \in I, j \in J} (f_{ij} - f_i f_j)^2 / f_i f_j $$

\item The term  $\| f_{IJ} - f_I f_J \|^2_{f_I f_J}$ is the $\chi^2$ metric
between the probability distribution $f_{IJ}$ and the product of marginal
distributions $f_I f_J$, with as centre of the metric the product
$f_I f_J$.  
\item Decomposing the moment of inertia of the cloud $N_J(I)$ -- or 
of $N_I(J)$ since both analyses are inherently related -- furnishes the 
principal axes of inertia, defined from a singular value decomposition.
\end{itemize}

\subsection*{Output: Cloud of Points Endowed with the Euclidean 
Metric in Factor Space}

\begin{itemize}

\item The $\chi^2$ distance with centre $f_J$ between observations $i$ and 
$i'$ is written as follows in two different notations: 

$$
d(i,i') = \| f^i_J - f^{i'}_J \|^2_{f_J} = \sum_j \frac{1}{f_j} 
\left( \frac{f_{ij}}{f_i} - \frac{f_{i'j}}{f_{i'}} \right)^2
$$

\item In the factor space this pairwise distance is identical.  The coordinate
system and the metric change.  

\item For factors indexed by $\alpha$ and for 
total dimensionality $N$ ($ N = \mbox{ min } \{ |I| - 1, |J| - 1 \}$;
the subtraction of 1 is since 
the $\chi^2$ distance is centred and  
hence there is a linear dependency which 
reduces the inherent dimensionality by 1) we have the projection of 
observation $i$ on the $\alpha$th factor, $F_\alpha$, given by 
$F_\alpha(i)$: 

\begin{equation}
d(i,i') = \sum_{\alpha = 1..N} \left( F_\alpha(i) - F_\alpha(i') \right)^2
\end{equation}

\end{itemize}

\begin{itemize}

\item In Correspondence Analysis the factors are ordered by decreasing 
moments of inertia.  
\item The factors are closely related, mathematically, 
in the decomposition of the overall cloud, 
$N_J(I)$ and $N_I(J)$, inertias.  
\item The eigenvalues associated with the 
factors, identically in the space of observations indexed by set $I$, 
and in the space of attributes indexed by set $J$, are given by the 
eigenvalues associated with the decomposition of the inertia.  
\item The 
decomposition of the inertia is a 
principal axis decomposition, which is arrived at through a singular
value decomposition.  
\end{itemize}

\subsection*{Supplementary Elements: Information Space Fusion}

\subsubsection*{Dual Spaces and Transition Formulas}

\begin{itemize}

\item 
$$ F_\alpha(i) = \lambda^{-\frac{1}{2}}_\alpha \sum_{j \in J}
f^i_j G_\alpha(j) \mbox{  for  } \alpha = 1, 2, \dots N; i \in I$$

$$ G_\alpha(j) = \lambda^{-\frac{1}{2}}_\alpha \sum_{i \in I} 
f^j_i F_\alpha(i) \mbox{  for  } \alpha = 1, 2, \dots N; j \in J$$

\item {\em Transition formulas}:  The coordinate of
element $i \in I$ is the barycentre of the coordinates of the elements
$j \in J$, with associated masses of value given by the coordinates of
$f^i_j$ of the profile $f^i_J$.  This is all to within the
$\lambda^{-\frac{1}{2}}_\alpha$ constant.

\item In the
output display, the barycentric principle comes into play: this allows
us to simultaneously view and interpret observations and attributes.

\end{itemize}

\subsubsection*{Supplementary Elements}

Overly-preponderant elements (i.e.\ row or column profiles), or
exceptional elements (e.g.\ a sex attribute, given other performance or
behavioural attributes) may be placed as supplementary elements.  This
means that they are given zero mass in the analysis, and their projections
are determined using the transition formulas.  This amounts to carrying
out a Correspondence Analysis first, without these elements, and then
projecting them into the factor space following the determination of all
properties of this space.

Here too we have a new approach to 
fusion of information spaces focusing the projection.

\subsection*{Hierarchical Clustering}

Background on hierarchical clustering in general, and the 
particular algorithm used here, can be found in \cite{murtagh85}.

Consider the projection of observation $i$ onto the set of 
all factors indexed by $\alpha$, $\{ F_\alpha(i) \}$ for all $\alpha$,
which defines the observation $i$ in the new coordinate frame.  
This new factor space is 
endowed with the (unweighted) Euclidean distance,  $d$.  
We seek a hierarchical clustering that takes into account the 
observation  sequence,
i.e.\ observation $i$ precedes observation $i'$ for all $i, i' \in I$.  
We use the linear
order on the observation set.  
\smallskip

\noindent
Agglomerative hierarchical clustering algorithm:  

\begin{enumerate}
\item Consider each observation in the sequence as constituting a 
singleton cluster.  
Determine the closest pair of adjacent observations, and define a cluster
from them.
\item Determine and merge 
the closest pair of adjacent clusters, $c_1$ and $c_2$, 
where closeness is defined by $d(c_1, c_2) = \mbox{ max } \{ d_{ii'}$
$\mbox{ such that } i \in c_1, i' \in c_2 \}$.  
\item Repeat the second step until only one cluster remains.
\end{enumerate}

This is a sequence-constrained complete link agglomeration criterion.  
The cluster proximity at each agglomeration is strictly non-decreasing.

\section*{Acknowledgements}

The media work is in collaboration with Adam Ganz and Stewart McKie,
Royal Holloway, University of London, Department of Media Arts.

\end{document}